 \documentclass[final,5p,times,authoryear]{elsarticle}
\usepackage{amssymb}
\usepackage{amsthm}
\usepackage{amsmath}
\usepackage{bm}
\usepackage[colorlinks=true]{hyperref}
\usepackage{subcaption}
\usepackage{float}
\usepackage[switch]{lineno}
\makeatletter
\def\ps@pprintTitle{%
  \let\@oddhead\@empty
  \let\@evenhead\@empty
  \let\@oddfoot\@empty
  \let\@evenfoot\@oddfoot
}
\makeatother

\usepackage[some]{background}
\SetBgScale{1}
\SetBgContents{\parbox{10cm}{%
  \Huge Draft:  \today\\[18cm]\rotatebox{180}{\Huge Draft:  \today}}}
\SetBgColor{red}
\SetBgAngle{270}
\SetBgOpacity{0.2}
\theoremstyle{definition}
\newtheorem{definition}{Definition}
\newtheorem{proposition}{Proposition}
\newtheorem{remark}{Remark}

\begin{document}

\begin{frontmatter}

% Command to add the watermark in the title page specifying the version 
%\BgThispage
%%%%%%



\title{Towards a Thermodynamical Deep-Learning-Vision-Based Flexible Robotic Cell for Circular Healthcare}


\author{Federico Zocco\corref{cor1}}
\author{Denis Sleath}
\author{and Shahin Rahimifard}

\address{Centre for Sustainable Manufacturing and Recycling Technologies, Wolfson School of Mechanical, Electrical and Manufacturing Engineering, Loughborough University, England, United Kingdom}
            
\cortext[cor1]{Corresponding author.\\ E-mail addresses: federico.zocco.fz@gmail.com (F. Zocco), d.m.sleath2@lboro.ac.uk (D. Sleath), s.rahimifard@lboro.ac.uk (S. Rahimifard).}

\begin{abstract}
The dependence on finite reserves of raw materials and the production of waste are two unsolved problems of the traditional linear economy. Healthcare, as a major sector of any nation, is currently facing them. Hence, in this paper, we report theoretical and practical advances of robotic reprocessing of small medical devices. Specifically, on the theory, we combine compartmental dynamical thermodynamics with the mechanics of robots to integrate robotics into a system-level perspective, and then, propose graph-based circularity indicators by leveraging our thermodynamic framework. Our thermodynamic framework is also a step forward in defining the theoretical foundations of circular material flow designs as it improves material flow analysis (MFA) by adding dynamical energy balances to the usual mass balances. On the practice, we report on the on-going design of a flexible robotic cell enabled by deep-learning vision for resources mapping and quantification, disassembly, and waste sorting of small medical devices.  % 149 < 150 words    
\end{abstract}



\end{frontmatter}

%\linenumbers

% For citations use: 
%       \citet{<label>} ==> Jones et al. [21]
%       \citep{<label>} ==> [21]
%

% If you have bibdatabase file and want bibtex to generate the
% bibitems, please use
%
%  \bibliographystyle{elsarticle-num-names} 
%  \bibliography{<your bibdatabase>}

% else use the following coding to input the bibitems directly in the
% TeX file.

\section{Introduction}
The OECD global material resources outlook to 2060 states that the global primary materials use is projected to almost double from 89 gigatonnes in 2017 to 167 gigatonnes in 2060 \citep{OECD}. The strongest growth in materials use is expected to occur in emerging and developing countries; even in the OECD countries the growth should be between 1\% to 2\% per year on average \citep{OECD}. Since most of the materials are finite resources, the transition to a more circular and resource efficient economy is essential to prevent future material supply shortages \citep{van2021circular}.  

The second benefit of implementing a more circular economy, i.e., an economy based on reducing, reusing, repairing, and recycling, is that waste and pollution rates will be reduced since products and materials are kept in use for longer while output flows from processes and households are recirculated within the economy \citep{EllenMacArthur}. For example, in healthcare, the NHS alone produces 156,000 tonnes of clinical waste each year \citep{NHS}, whose recovery must be both economically viable and free from contamination risks. According to \cite{shrank2019waste}, the implementation of effective measures to eliminate wastes represents an opportunity to reduce the US healthcare expenditure, specifically it was estimated a potential reduction of 25\% in the total cost of waste. In 2017, the China's ban on its import of most plastic waste has also motivated the developed countries to implement reprocessing approaches locally \citep{wen2021china}.

To enhance a domestic implementation of circular material flows in healthcare, in this paper we make the following main contributions:         
\begin{itemize}
\item{Since autonomous reprocessing systems can reduce both costs and contamination risks, we report on the on-going development of a \emph{flexible robotic cell for reprocessing small medical devices} such as glucose meters and inhalers. Minor layout changes can set the cell for either waste sorting, disassembly, or resources mapping and quantification (see Sections \ref{subsec:DesignOfCell} and \ref{subsec:cellPerformance}).}
\item{So far, the theory of robot design and control is not integrated into the design of whole supply and recovery chains. To address this gap, we \emph{integrate the standard form of robot dynamics into an holistic framework} based on compartmental dynamical thermodynamics. The integration has been possible thanks to the generality of thermodynamics and its link with Lagrangian mechanics (see Section \ref{sec:RobAndSysThinking}).}
\item{Our holistic framework assumes multiple thermodynamic compartments connected by material flows. This naturally led to modeling the robotic cell as a directed graph, and subsequently, to the derivation of \emph{indicators measuring the performance of the cell in terms of material flow circularity} (see Sections \ref{subsec:CircInd} and \ref{subsec:exampleIndicators}).}
\item{This paper coherently intersects dynamical thermodynamics, robotics theory, deep-learning-based computer vision, and graph theory to \emph{enhance both the theoretical foundations and the practical implementation of a circular economy}. Software and a demo video are publicly available\footnotemark{}. \footnotetext{\url{https://github.com/fedezocco/ThermoVisMedRob}}}
\end{itemize}

% A few paragraphs on related work:
Related work is as follows. Robotic disassembly to recover rare Earth materials from electronic components of electric vehicles was proposed by \cite{li2018robotic}, while disassembly sequence planning was addressed by \cite{wang2021energy,laili2021robotic}. An extensive book on disassembly is \cite{laili2022optimisation}. \cite{kiyokawa2022challenges} recently reviewed the most advanced robot functionalities discussing how to extend the flexibility of traditional waste sorting facilities, which target only a limited range of items. As it will be shown, we also seek flexibility in our robotic cell. Waste sorting of clinical waste was proposed in 2020 by \cite{devi2020automatic}, whereas \cite{du2022efficient} and \cite{raptopoulos2020robotic} focused on textiles and urban waste, respectively. 

One of the latest improvements in robot functionalities is brought by the recent breakthroughs in machine learning and consists of more accurate and generalizing vision models for visual servoing, i.e., for controlling the robot motion via real-time processing of images acquired by cameras \citep{bateux2018training,ribeiro2021real,hay2023noise,zhang2015towards}. Effective visual servoing could leverage the systems proposed by \cite{zocco2023visual,brogan2021deep,koskinopoulou2021robotic,kumar2021artificial} for general and flexible material, object and part recognition. As we will show, advanced vision is at the core of our robotic cell. 

To date, MFA is a key methodology to assess material flow circularity; essentially, it is based on the principle of mass conservation and on the analysis of real material stock and flow data \citep{cullen2022material}. As it was shown in \cite{zocco2023thermodynamical,zocco2022circularity} and further in this paper, our thermodynamic framework enhances MFA in two aspects: first, it adds dynamical energy balances \citep{haddad2019dynamical} to the usual mass balances; and second, it is underpinned by continuous- and discrete-time dynamical systems rather than by data-analysis, and hence, it is more accurate in simulating what-if scenarios, in which real data are missing (see \cite{MITlectures} and \cite{haddad2008nonlinear} for an introduction and an advanced book on dynamical systems, respectively). The lack of real data is very common in circular economy research since circular material flows are still rare in reality.

Throughout the paper, vectors and matrices are indicated with bold lower-case and upper-case letters, respectively, while sets are indicated with upper-case calligraphic letters. 

The paper is structured as follows: Section \ref{sec:methods} details the robotic cell design, the thermodynamic framework, and the circularity indicators; then, Section \ref{sec:results} shows and discusses the results and future work; finally, Section \ref{sec:concl} concludes.

\section{Methods}\label{sec:methods}
This section begins by integrating the theory of robot mechanics into the systemic modeling framework of thermodynamical material networks (TMNs) \citep{zocco2023thermodynamical}, which is required to address system-level questions such as those arising in circular economy. Then, it describes the robotic cell we are developing and proposes two circularity indicators to assess its performance.
  
\subsection{Integration of Robotics into a Systemic Framework}\label{sec:RobAndSysThinking}
\subsubsection{Principles of TMNs and Circularity}
The definition of a TMN results from the combination of graph theory \citep{bondy1976graph} and compartmental dynamical thermodynamics \citep{haddad2019dynamical}: the former is a well-established approach for analyzing networks (e.g., hydraulic networks \citep{sitzenfrei2023graph}, electrical networks \citep{freitas2020stochastic}, and epidemic networks \citep{gomez2023new}), while the latter is a formalism that generalizes the mature design technique of water and thermal systems \citep{kaminski2017introduction}. To note that, in this paper, we go straight to the definition of the discrete-time mass-flow matrix, which assumes familiarity with the preliminary definitions covered by \cite{zocco2023thermodynamical,zocco2022circularity}.   

\begin{definition}[Discrete-time mass-flow matrix]\label{def:discreteMFmat}
Given a TMN
\begin{equation}\label{def:TMNset}
\begin{gathered}
\mathcal{N} = \left\{c^1_{1,1}, \dots, c^{k_v}_{k_v,k_v}, \dots, c^{n_v}_{n_v,n_v}, \right. \\ 
\left. c^{n_v+1}_{i_{n_v+1},j_{n_v+1}}, \dots, c^{n_v+k_a}_{i_{n_v+k_a},j_{n_v+k_a}}, \dots, c^{n_c}_{i_{n_c},j_{n_c}}\right\}, 
\end{gathered}
\end{equation}
the \emph{discrete-time mass-flow matrix} $\bm{\Gamma}(\mathcal{N}; n)$ associated with the network (\ref{def:TMNset}) is
\begin{equation}\label{eq:gammaDef}
\begin{split}
\bm{\Gamma}(\mathcal{N}; n) & =
\begin{bmatrix}
\gamma_{1,1}(n+1) & \dots & \gamma_{1,n_v}(n+1) \\
    \vdots & \ddots & \vdots \\
\gamma_{n_v,1}(n+1) &  \dots & \gamma_{n_v,n_v}(n+1) 
\end{bmatrix}
\\ 
& = 
\begin{bmatrix}
m_1(n+1) & m_{1,2}(n+1) & \dots & m_{1,n_v}(n+1) \\
m_{2,1}(n+1) & m_2(n+1) & \dots & m_{2,n_v}(n+1) \\
\vdots & \vdots & \ddots & \vdots \\
m_{n_v,1}(n+1) & m_{n_v,2}(n+1) & \dots & m_{n_v}(n+1) 
\end{bmatrix}
\end{split} 
\end{equation}
where $n \in \overline{\mathbb{Z}}_+$, $\bm{\Gamma}(\mathcal{N}; n) \in \mathbb{R}^{n_v \, \times \, n_v}$, $\overline{\mathbb{Z}}_+$ is the set of non-negative integers, $n_v$ is the number of vertices, $n_c$ is the total number of compartments, the entries along the diagonal are the weights of the vertex-compartments $c^{k_v}_{k_v,k_v} \in \mathcal{R}$ (i.e., mass stocks), with $\mathcal{R} \subseteq \mathcal{N}$ the subset of compartments that \emph{store}, \emph{transform}, or \emph{use} the target material, and whose off-diagonal entries are the weights of the arc-compartments $c^k_{i,j} \in \mathcal{T}$ (i.e., masses that move between the vertices), with $\mathcal{T} \subseteq \mathcal{N}$ the subset of compartments that \emph{move} the target material between the compartments belonging to $\mathcal{R} \subseteq \mathcal{N}$.
\end{definition}

Conceiving the material life-cycle as a sequence of thermodynamic compartments that transport and transform the material leads to the following definition of circularity.
\begin{definition}[\cite{zocco2023thermodynamical}]\label{def:ThermoCircular}
The flow of a material $\beta$ is \emph{thermodynamically} circular if there exists an ordered sequence of thermodynamic compartments 
\begin{equation}
\phi = \left(c_{1,1}^1, \dots, c_{i,j}^k, \dots, c_{1,1}^1\right)
\end{equation}
processing $\beta$ which begins and ends in $c_{1,1}^1$. More generally, the flow of the material set $\mathcal{B} = \{\beta_1, \dots, \beta_k, \dots, \beta_{n_{\beta}}\}$ is thermodynamically circular if there exists an ordered sequence $\phi$ processing $\mathcal{B}$. 
\end{definition}

\subsubsection{Robot as a Thermodynamic Compartment}\label{sub:RobotAsThermoComp}
The previous section has defined a systemic modeling framework in which the material flow results from a network of thermodynamic compartments connected by the exchange of materials. Therefore, if we demonstrate that a robot can be seen as a thermodynamic compartment of such a network, we achieve the integration of robotics theory into this systemic framework so that we can design industrial robotic operations as part of supply and recovery chains (i.e., networks) instead of stand alone. A systemic framework is essential to address circular economy questions as they are systemic in their nature. Such an integration of robotics theory is achieved with the following proposition.
\begin{proposition}\label{prop:MainProp}
The standard form of robot dynamics (see Ch. 7 in \cite{SicilianoBook})
\begin{equation}
\begin{gathered}\label{eq:standardFormDynamics}
\sum_{j=1}^\alpha b_{ij}(\bm{q})\ddot{q}_j + \sum_{j=1}^\alpha \sum_{k=1}^\alpha h_{ijk}(\bm{q})\dot{q}_k\dot{q}_j \\ + g_i(\bm{q}) = \xi_i, \quad i = 1, \dots, \alpha,
\end{gathered}
\end{equation}
can be derived from the dynamical form of the first principle of thermodynamics 
\begin{equation}\label{eq:dynFromFirstThermo}
\frac{\text{d}E}{\text{d}{t}} = \dot{Q} - \dot{W},
\end{equation} 
where $E$ is the total energy of the robotic manipulator, $\dot{Q}$ is the heat flow exchanged between the manipulator and the surroundings, $\dot{W}$ is the work flow exchanged between the manipulator and the surroundings, $\alpha$ is the number of rigid links of the robotic manipulator, $\bm{q}$ is the vector of generalized coordinates, $q_j$ is the $j$-th element of $\bm{q}$, $\bm{\xi}$ is the vector of generalized forces, and where $b_{ij}(\bm{q})$, $h_{ijk}(\bm{q})$, and $g_i(\bm{q})$ are coefficients that take into account the gravity, the Coriolis effect, the centrifugal effect, and the inertias of the links.   
\end{proposition}
It follows from Proposition \ref{prop:MainProp} that a robot can be seen as a thermodynamic compartment, and hence, its modeling and control can be embedded into the design of a TMN \citep{zocco2023thermodynamical} to address the system-level questions concerning material flow circularity.
\begin{proof}
Consider the dynamical form of the first principle of thermodynamics (\ref{eq:dynFromFirstThermo}) with the total energy contributions developed as $E = K + U + P$, where $U$ is the internal energy, $K$ is the kinetic energy, and $P$ is the potential energy of the system under study. Since $U$ and the heat flow $\dot{Q}$ are neglected in solid mechanics, (\ref{eq:dynFromFirstThermo}) reduces to
\begin{equation}\label{eq:dynFromFirstThermoSolMech}
\frac{\text{d}}{\text{d}{t}}(K + P) = - \dot{W}.
\end{equation}  
Taking the partial derivative on both sides of (\ref{eq:dynFromFirstThermoSolMech}) with respect to $\bm{\dot{q}}$ and transposing the resulting equation yields 
\begin{equation}\label{eq:powerBalance-2}
\frac{\text{d}}{\text{d}{t}}\left[\frac{\partial (K + P)}{\partial \dot{\bm{q}}}\right]^\top = - \left(\frac{\partial \dot{W}}{\partial \dot{\bm{q}}}\right)^\top.
\end{equation}
Recalling the Lagrangian function
\begin{equation}\label{eq:LagrFunc}
L = K - P
\end{equation}
and the fact that $\partial P / \partial \dot{\bm{q}} = 0$, (\ref{eq:powerBalance-2}) can be written as
\begin{equation}\label{eq:powerBalance-3}
\frac{\text{d}}{\text{d}{t}}\left(\frac{\partial L}{\partial \dot{\bm{q}}}\right)^\top = - \left(\frac{\partial \dot{W}}{\partial \dot{\bm{q}}}\right)^\top.
\end{equation} 
Note now that the left-hand side of the formulation of the Lagrange's equations of motion
\begin{equation}\label{eq:LagrangianEquations}
\frac{\text{d}}{\text{d}{t}}\left(\frac{\partial L}{\partial \dot{\bm{q}}}\right)^\top = \bm{\xi} + \left(\frac{\partial L}{\partial \bm{q}} \right)^\top
\end{equation}
is equal to the left-hand side of (\ref{eq:powerBalance-3}). Hence, the right-hand sides of (\ref{eq:LagrangianEquations}) and (\ref{eq:powerBalance-3}) are also the same. Thus, (\ref{eq:powerBalance-3}) can be written as (\ref{eq:LagrangianEquations}), and hence, the Lagrange's equations of motion (\ref{eq:LagrangianEquations}) can be seen as a derivation of the power balance (\ref{eq:dynFromFirstThermo}). To complete the proof, we will now show that (\ref{eq:LagrangianEquations}) can be written as the standard form of robot dynamics (\ref{eq:standardFormDynamics}).        

Consider a manipulator with $\alpha$ rigid links. The total kinetic energy is the sum of the kinetic energy of all the links and all the motors (see Ch. 7 in \cite{SicilianoBook}), that is,
\begin{equation}\label{eq:kinEnergyLinkAndMotor}
K = \sum_{i=1}^\alpha (K_{\text{l}_i} + K_{\text{m}_i}),  
\end{equation}
where $K_{\text{l}_i}$ is the kinetic energy of the $i$-th link and $K_{\text{m}_i}$ is the kinetic energy of the motor actuating Joint $i$. The kinetic energy contribution of Link $i$ is
\begin{equation}\label{eq:kinEnergy}
K_{\text{l}_i} = \frac{1}{2} \int _{V_{\text{l}_i}} \dot{\bm{p}}^{*\top}_i \dot{\bm{p}}^*_i \rho \text{d}V,
\end{equation}    
where $\dot{\bm{p}}^*_i$ denotes the linear velocity vector and $\rho$ is the density of the elementary particle of volume $\text{d}V$; $V_{\text{l}_i}$ is the volume of Link $i$.
Be $\bm{p}_i^*$ the position vector of the elementary particle and be $\bm{p}_{\text{l}_i}$ the position vector of the link centre of mass expressed in the base frame. Hence, we can write 
\begin{equation}
\bm{r}_i = \bm{p}^*_i - \bm{p}_{\text{l}_i} 
\end{equation}
and, thus, the link particle velocity can be written as   
\begin{equation}\label{eq:VelCofM}
\dot{\bm{p}}^*_i = \dot{\bm{p}}_{\text{l}_i} + \bm{\omega}_i \times \bm{r}_i = \dot{\bm{p}}_{\text{l}_i} + \bm{S}(\bm{\omega}_i)\bm{r}_i, 
\end{equation}
with $\dot{\bm{p}}_{\text{l}_i}$ the linear velocity of the center of mass, $\bm{\omega}_i$ the angular velocity of the $i$-th link, and $\bm{S}(\cdot)$ a matrix operator (see Ch. 7 in \cite{SicilianoBook}). By substituting (\ref{eq:VelCofM}) into (\ref{eq:kinEnergy}), we have that
\begin{equation}\label{eq:kinEnergy2}
K_{\text{l}_i} = \frac{1}{2} \int _{V_{\text{l}_i}} \left(\dot{\bm{p}}_{\text{l}_i} + \bm{S}(\bm{\omega}_i)\bm{r}_i\right)^\top \left( \dot{\bm{p}}_{\text{l}_i} + \bm{S}(\bm{\omega}_i)\bm{r}_i\right) \rho \text{d}V,
\end{equation}
which contains three contributions to the velocity: the translational term, the rotational term, and the mutual term. By introducing the Jacobian columns relative to the joint velocities up to the Link $i$ with the matrices $J^{(\text{l}_i)}_{\text{P}}$ and $J^{(\text{l}_i)}_{\text{O}}$, the kinetic energy of Link $i$ in (\ref{eq:kinEnergy2}) can be written as a function of $\bm{\bm{q}}$ as
\begin{equation}\label{eq:KinEnergyLinkFinal}
K_{\text{l}_i} = \frac{1}{2} m_{\text{l}_i} \dot{\bm{q}}^\top J^{(\text{l}_i)\top}_{\text{P}} J^{(\text{l}_i)}_{\text{P}} \dot{\bm{q}} + \frac{1}{2} \dot{\bm{q}}^\top J^{(\text{l}_i)\top}_{\text{O}} \bm{R}_i \bm{I}^i_{\text{l}_i} \bm{R}_i^\top J^{(\text{l}_i)}_{\text{O}} \dot{\bm{q}},
\end{equation} 
where $m_{\text{l}_i}$ is the link mass, $\bm{R}_i$ is the rotation matrix from Link $i$ frame to the base frame, and $\bm{I}^i_{\text{l}_i}$ is the inertia tensor relative to the centre of mass of Link $i$ when expressed in the link frame (see Ch. 7 in \cite{SicilianoBook}). The kinetic energy contribution of the motor of the $i$-th joint can be computed in an analogous way to that of the link.

Finally, by summing the contributions of the single links (\ref{eq:KinEnergyLinkFinal}) and motors as in (\ref{eq:kinEnergyLinkAndMotor}), the total kinetic energy of the manipulator is given by the quadratic form 
\begin{equation}\label{eq:KinEnergyFinal}
K(\bm{q}, \dot{\bm{q}}) = \frac{1}{2} \sum_{i=1}^\alpha \sum_{j=1}^\alpha b_{ij}(\bm{q})\dot{q_i} \dot{q}_j = \frac{1}{2} \dot{\bm{q}}^\top \bm{B}(\bm{q}) \dot{\bm{q}},
\end{equation}      
where $B(\bm{q})$ is the inertia matrix (see Ch. 7 in \cite{SicilianoBook}). 

Let us now calculate the potential energy $P$ of the manipulator as a function of the generalized coordinates. On the assumption of rigid links, the only conservative force is gravitational whose potential energy is given by 
\begin{equation}\label{eq:PotEnergyFinal}
P(\bm{q}) = - \sum_{i = 1}^\alpha (m_{\text{l}_i} \bm{g}_0^\top \bm{p}_{\text{l}_i} + m_{\text{m}_i} \bm{g}_0^\top \bm{p}_{\text{m}_i}),
\end{equation} 
where $\bm{g}_0$ is the gravity acceleration vector in the base frame, $m_{\text{m}_i}$ is the mass of the $i$-th motor, and $\bm{p}_{\text{m}_i}$ is the position of the centre of mass of the motor in the base frame.

Let us now substitute the kinetic energy (\ref{eq:KinEnergyFinal}) and the potential energy (\ref{eq:PotEnergyFinal}) as functions of the generalized coordinates and velocities into the Lagrangian function (\ref{eq:LagrFunc}) to get
\begin{equation}\label{eq:LagrangianWithq}
L(\bm{q}, \dot{\bm{q}}) = K(\bm{q}, \dot{\bm{q}}) - P(\bm{q}).
\end{equation}    
Inserting (\ref{eq:LagrangianWithq}) into the Lagrange's equations of motion (\ref{eq:LagrangianEquations}) and taking the required derivatives (note that $P$ does not depend on $\dot{\bm{q}}$) yields the standard form of robot dynamics (\ref{eq:standardFormDynamics}) (see Ch. 7 in \cite{SicilianoBook}). Therefore, (\ref{eq:standardFormDynamics}) is a derivation of (\ref{eq:LagrangianEquations}), and hence, it is also a derivation of (\ref{eq:dynFromFirstThermo}).       
\end{proof}

\subsection{Design of Flexible Robotic Cell}\label{subsec:DesignOfCell}
\begin{figure*}
\begin{subfigure}{0.203\textwidth}
  \centering
  \includegraphics[width=.99\linewidth]{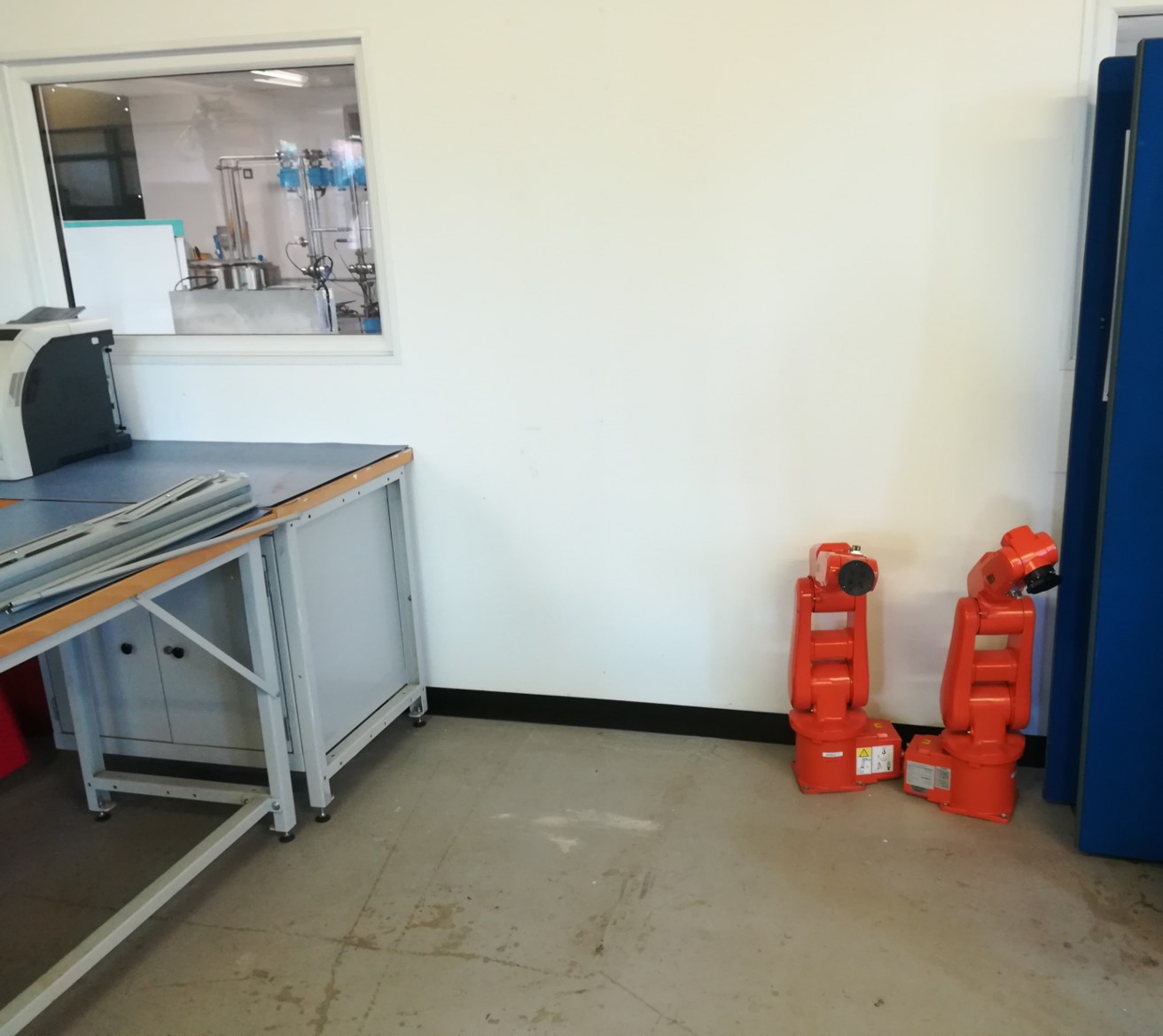}
  \caption{Starting point}
  \label{fig:initial}
  \vspace{2mm}
\end{subfigure}%
\begin{subfigure}{0.24\textwidth}
  \centering
  \includegraphics[width=.99\linewidth]{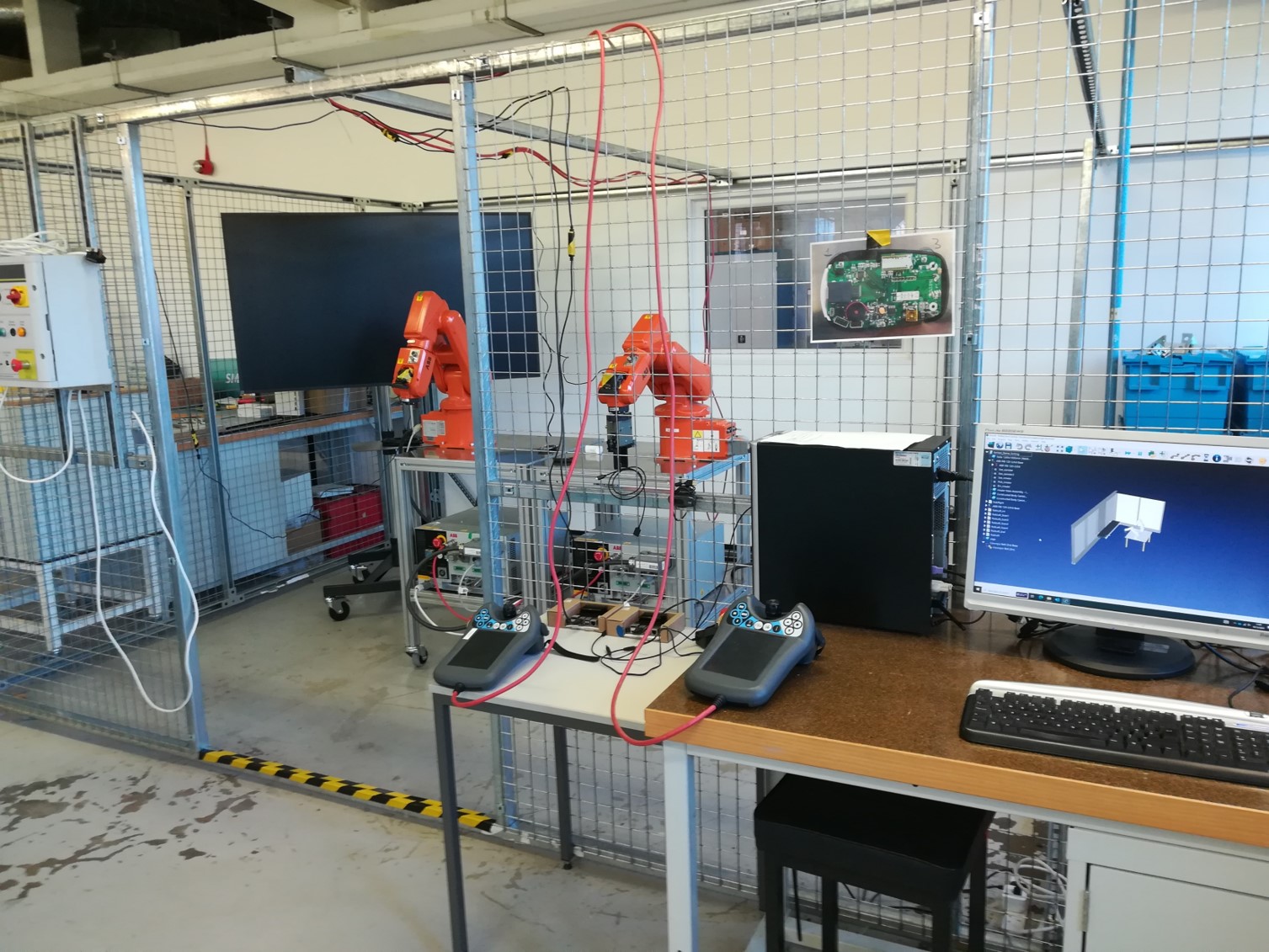}
  \caption{Now (as of January 2024)}
  \label{fig:global}
  \vspace{2mm}
\end{subfigure}
\begin{subfigure}{0.24\textwidth}
  \centering
  \includegraphics[width=.99\linewidth]{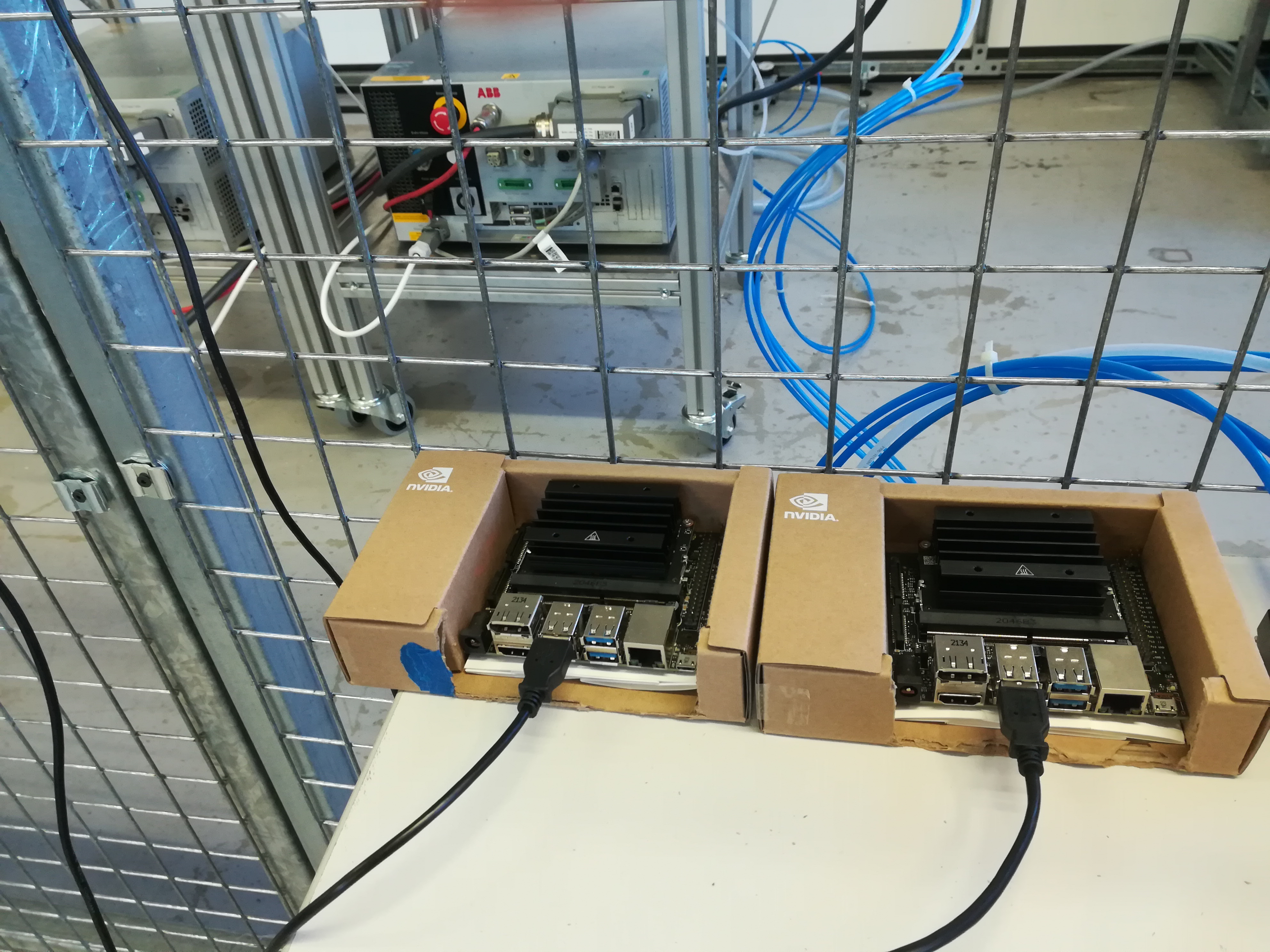}
  \caption{AI microprocessors}
  \label{fig:microproc}
  \vspace{2mm}
\end{subfigure}
\begin{subfigure}{0.24\textwidth}
  \centering
  \includegraphics[width=.99\linewidth]{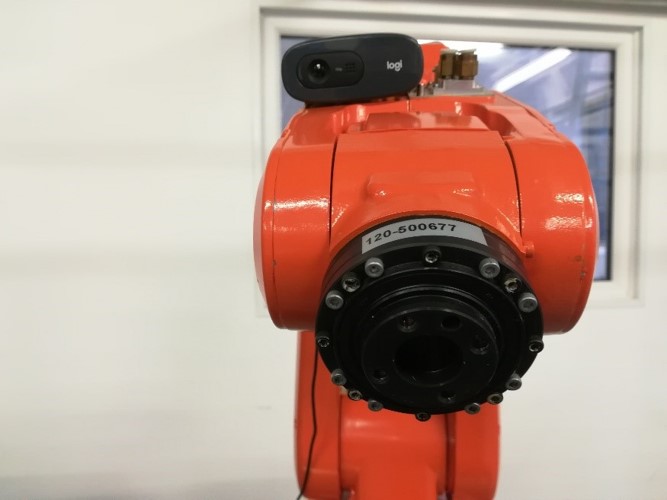}
  \caption{Robot webcam}
  \label{fig:camera}
  \vspace{2mm}
\end{subfigure} \\ 
\begin{subfigure}{0.26\textwidth}
  \centering
  \includegraphics[width=.99\linewidth]{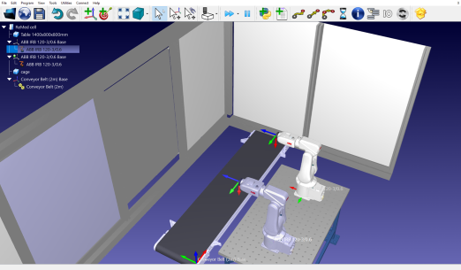}
  \caption{Cell simulator}
  \label{fig:simulator}
  %\vspace{5mm}
\end{subfigure}%
\begin{subfigure}{0.22\textwidth}
  \centering
  \includegraphics[width=.99\linewidth]{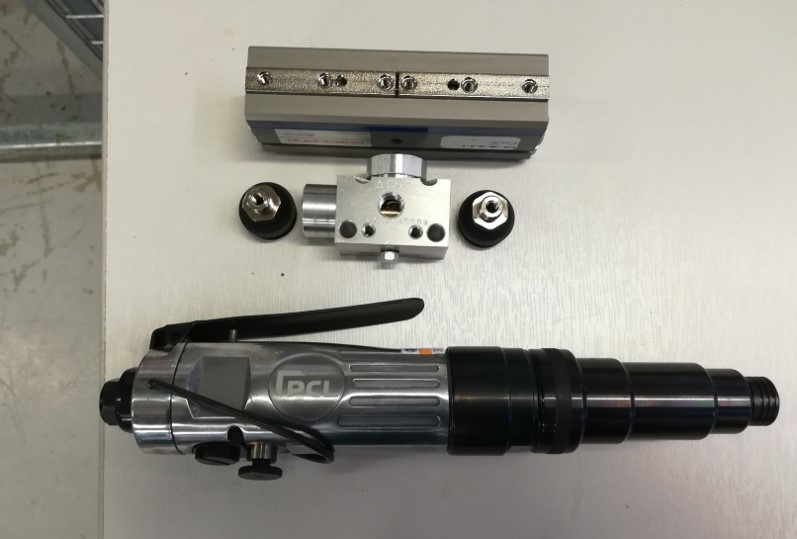}
  \caption{Robot tools}
  \label{fig:tools}
  %\vspace{5mm}
\end{subfigure}
\begin{subfigure}{0.20\textwidth}
  \centering
  \includegraphics[width=.99\linewidth]{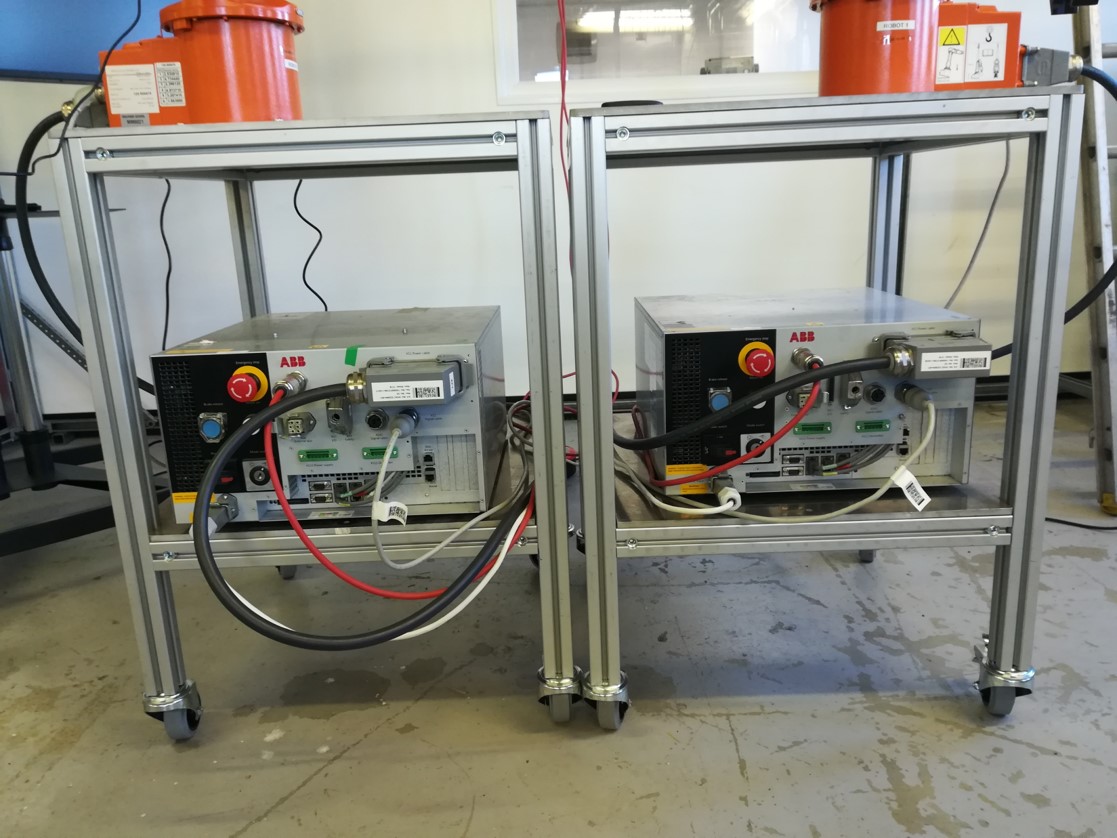}
  \caption{Castors}
  \label{fig:castors}
  %\vspace{5mm}
\end{subfigure}
\begin{subfigure}{0.30\textwidth}
  \centering
  \includegraphics[width=.99\linewidth]{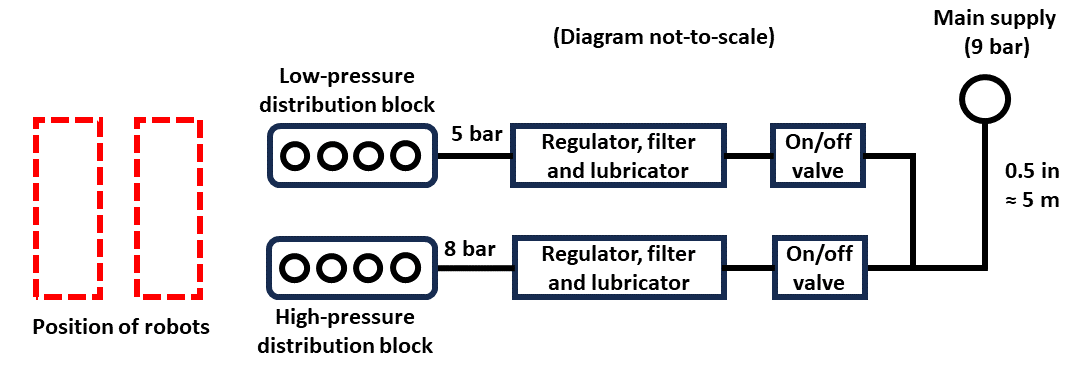}
  \caption{Pneumatic system}
  \label{fig:pneumatics}
  %\vspace{5mm}
\end{subfigure} 
\caption{Main components of the flexible robotic cell.}
\label{fig:CellComponents}
\end{figure*}

The main components of the flexible robotic cell we designed for reprocessing medical devices are shown in Fig. \ref{fig:CellComponents}. A quick overview of our progress is visible by comparing Fig. \ref{fig:initial} and Fig. \ref{fig:global}, which depict the initial and the current situation (as of January 2024), respectively. The cell occupies approximately an area of 5 $\times$ 2.5 m and it contains two ABB IRB 120 manipulators whose payload is 3 kg and whose reach is approximately 1.10 m (including the tool). The size of the robots has been chosen considering that the weight of the medical devices of our interest is less than 1.5 kg. These light robots are relatively easy to move to reconfigure the cell layout according to our research needs. The whole area is delimited by a safety cage since the ABB IRB 120 robot is not suitable to operate next to a human operator. A simulator of the robotic cell runs on the desktop beside the cage (bottom right of Fig. \ref{fig:global} and Fig. \ref{fig:simulator}) to check for collisions before operating the real robots and also to generate the robot programs. The simulator is developed on \cite{RoboDK}. Each robot has a Logitech C270 webcam attached in proximity of its flange as in Fig. \ref{fig:camera}. Each webcam is connected to an Nvidia Jetson Nano microprocessor running deep neural networks for real-time image processing (Fig. \ref{fig:microproc}). Each robot is supported by a frame that contains the controller (model ABB IRC5) and it is installed on braked castors for cell reconfigurability (Fig. \ref{fig:castors}). For example, waste sorting could be performed with the robots standing on the opposite sides of a conveyor belt, whereas disassembly could be performed with the robots facing each other.   
The robot tools are in Fig. \ref{fig:tools}, they are pneumatic and a diagram of the air system is shown in Fig. \ref{fig:pneumatics}. Specifically, we have a vacuum gripper for quick and easy pick-and-place, a parallel gripper for more complex manipulations, a grinder and a screwdriver for semi-destructive and non-destructive disassembly operations, respectively.

The Nvidia Jetson Nano microprocessor is a developer kit optimized for AI applications (Fig. \ref{fig:microproc}) \citep{DLIstartedNano}. We trained a ResNet-34 \citep{he2016deep} on each microprocessor with batch size of 8 and the Adam optimizer \citep{kingma2014adam}. One neural model was trained for 4 epochs on 266 images collected from the robot webcab (Fig. \ref{fig:camera}) to classify syringes, glucose meters, and inhalers for waste sorting operations. The other model was trained for 10 epochs on 435 samples to track the position of the 3 screws that secure the printed circuit board (PCB) of a glucose meter to the plastic case. The application of this second vision system is screw removal either via a grinder (semi-destructive disassembly) or via a screwdriver (non-destructive disassembly).

\subsection{Circularity Indicators of Robotic Cell}\label{subsec:CircInd}
\begin{figure}[H]
    \centering
    \includegraphics[width=0.4\textwidth]{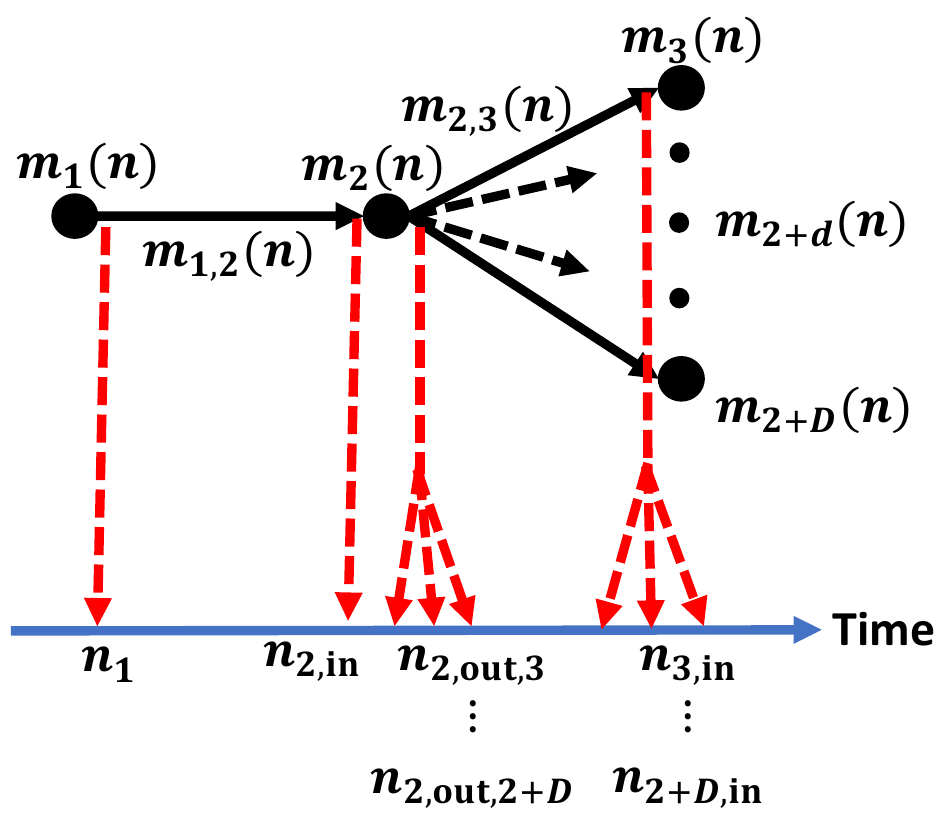}
    \caption{At the top, the mass-flow digraph of the robotic cell. The digraph is applicable to both disassembly and waste sorting scenarios. The red arrows indicate the corresponding time at which the mass leaves or enters a vertex-compartment.}
    \label{fig:CellGraph}
\end{figure}

To define the circularity indicators of the cell, first we model the cell as a mass-flow digraph $M(\mathcal{N})$, then we define the mass-flow matrix $\bm{\Gamma}(\mathcal{N}; n)$, and finally derive the indicators from the latter. The mass-flow digraph of the cell is shown in Fig. \ref{fig:CellGraph} and it is valid for both disassembly and waste sorting settings.
Specifically, in the case of disassembly, the digraph depicts the following scenario:  $m_1(n)$ is the mass of a stock of products to be disassembled, $m_{1,2}(n)$ is the mass moved to the disassembly cell, $m_2(n)$ is the mass being disassembled, $m_{2,3}(n)$ is the mass of the materials moved to the first bin (case of $d = 1$), $m_{2,4}(n)$ is the mass of the materials moved to the second bin (case of $d = 2$), and $m_{2,2+d}(n)$ is the mass of the materials moved to the $d$-th bin. The total number of bins is $D$. Hence, $n_v = 2+D$, the number of arcs is $n_a = 1 + D$, and $n_c = 3 + 2D$. The mass leaves the first vertex (i.e., the stock of products) at time $n_1$, it enters the second vertex (i.e., the disassembly line) at time $n_{2,\text{in}}$, it leaves the second vertex at time $n_{2,\text{out},2+d}$ to enter the $d$-th bin, and it enters the $d$-th bin at time $n_{2+d,\text{in}}$. In contrast, in the case of waste sorting, the same digraph depicts the following scenario: $m_1(n)$ is a stock of waste to be sorted, $m_{1,2}(n)$ is the fraction of waste sent to the sorting cell, $m_2(n)$ is the mass inside the sorting cell, $m_{2,2+d}(n)$ is the mass of the materials moved to the $d$-th bin.              

The TMN corresponding to the cell digraph (Fig. \ref{fig:CellGraph}) is 
\begin{equation}
\begin{gathered}
\mathcal{N}^{\text{c}} = \left\{c_{1,1}^1, c_{2,2}^2, \dots, c_{2+d,2+d}^{2+d}, \dots, c_{2+D,2+D}^{2+D}, c_{1,2}^{3+D}, \dots,  \right. \\ \left. c_{2,2+d}^{3+D+d}, \dots, c_{2,2+D}^{3+2D} \right\}, \quad d = 1, \dots, D,
\end{gathered}
\end{equation} 
while the mass-flow matrix (Definition \ref{def:discreteMFmat}) becomes
\begin{equation}\label{eq:GammaOfCell}
\bm{\Gamma}^{\text{c}}(\mathcal{N}; n) =
\begin{bmatrix}
m_1(n+1) & m_{1,2}(n+1) & \dots & 0 \\
0 & m_2(n+1) & \dots & m_{2,2+D}(n+1) \\
\vdots & \vdots & \ddots \\
0 & 0 & \dots & m_{2+D}(n+1)
\end{bmatrix},
\end{equation}
where $\bm{\Gamma}^{\text{c}}(\mathcal{N}; n) \in \mathbb{R}^{(2 + D) \, \times \, (2 + D)}$ and $n \in \overline{\mathbb{Z}}_+$. 

Let us now specify the entries of (\ref{eq:GammaOfCell}). The off-diagonal entries can be written as first order difference equations, that is,
\begin{equation}\label{eq:FlowFrom1to2}
m_{1,2}(n + 1) - m_{1,2}(n) = \overline{m}_{1,2} \left(\delta_{n_1}(n) - \delta_{n_{2,\text{in}}}(n)\right) 
\end{equation}
and
\begin{equation}
\begin{gathered}
m_{2,2+d}(n + 1) - m_{2,2+d}(n) = \overline{m}_{2,2+d} \left(\delta_{n_{2,\text{out},2+d}} \right. \\ \left. - \delta_{n_{2+d,\text{in}}} \right), \quad d = 1, \dots, D, 
\end{gathered}
\end{equation}
where
\begin{equation}\label{eq:defOfImpulse}
\delta_{n_1}(n) =
\begin{cases}
      0, & \, n \neq n_1 \\
      1, & \, n = n_1,
\end{cases}
\end{equation}
which is the Kronecker delta.
The definitions of the other Kronecker deltas are analogous to (\ref{eq:defOfImpulse}). The parameter $\overline{m}_{i,j}$ is the mass moving between vertex $i$ and vertex $j$ during the time window specified by the multiplying Kronecker deltas, e.g., for $\overline{m}_{1,2}$ in (\ref{eq:FlowFrom1to2}) the time window starts in $n_1$ and ends in $n_{2,\text{in}}$.   

The entries along the main diagonal follow from the imposition of the principle of mass conservation in discrete time, that is,
\begin{equation}
m_1(n + 1) - m_1(n) = - \overline{m}_{1,2} \delta_{n_1}(n),
\end{equation}
\begin{equation}
\begin{gathered}
m_2(n + 1) - m_2(n) = \overline{m}_{1,2} \delta_{n_{2,\text{in}}}(n) \\ - [\overline{m}_{2,3} \delta_{n_{2,\text{out},3}}(n) + \dots + \overline{m}_{2,2+d} \delta_{n_{2,\text{out},2+d}}(n) \\ + \dots + \overline{m}_{2,2+D} \delta_{n_{2,\text{out},2+D}}(n)]_{d = 2, \dots ,D-1}\\
\end{gathered} 
\end{equation}
and
\begin{equation}
m_{2+d}(n + 1) - m_{2+d}(n) = \overline{m}_{2,2+d} \delta_{n_{2+d,\text{in}}}(n), \quad d = 1, \dots, D. 
\end{equation}
As the cell digraph is a closed system, from the principle of conservation of mass it follows that
\begin{equation}
\sum_{i=1,j = 1}^{2+D} \gamma^{\text{c}}_{i,j}(n) = \text{const.} \quad \forall n,
\end{equation}
where $\gamma^{\text{c}}_{i,j}$ is the $(i,j)$ entry of $\bm{\Gamma}^{\text{c}}$.  

The mass-flow matrix (\ref{eq:GammaOfCell}) contains useful information related to the material circularity: the stock masses are along the main diagonal, the mass flows are off-diagonal, the times $n_1$, $n_{2,\text{in}}$, etc. at which a certain mass enters or leaves a stock is expressed by each entry, and it also contains the information about the layout of the cell, e.g., the number of rows of $\bm{\Gamma}^{\text{c}}(\mathcal{N}; n)$ depends on the number of bins $D$, while the entry $\gamma_{i,j}(n)$ indicates the material flow from the vertex $i$ to the vertex $j$; if $\gamma_{i,j}(n) = 0$, the two vertices are not exchanging mass directly (for example, the stock $m_1$ and the first bin $m_3$).     

We now define two circularity indicators for the robotic cell using the mass-flow matrix (\ref{eq:GammaOfCell}). The first indicator is the \emph{separation rate} $r_s$ defined as
\begin{equation}\label{eq:sepRate}
r_s \triangleq \text{Number of rows of }\, \bm{\Gamma}^{\text{c}}(\mathcal{N}; n),
\end{equation}  
which quantifies to what extent the cell separates the input flow $m_{1,2}(n)$ into different flows $m_{2,2+d}(n)$ to facilitate the recovery of the materials or the parts. An increase in the number of bins $D$ yields an increase of $r_s$ since $r_s = 2+ D$. The second circularity indicator is the \emph{separation time} $t_s$ defined as
\begin{equation}\label{eq:sepTime}
t_s \triangleq \max_{\substack{d \\ 1 \leq d \leq D}} \left(n_{2,\text{out},2+d} - n_{2,\text{in}}\right)T,
\end{equation}  
where $T$ is the sample time at which the dynamics of the system is recorded. In words, $t_s$ is the time that passes from when $\overline{m}_{1,2}$ enters the robotic cell (i.e., vertex 2) to when the \emph{last} mass leaves the robotic cell to enter the designated $d$-th bin (i.e., vertex $2+d$).  

\begin{remark}
The indicators $r_s$ and $t_s$ could have been defined without introducing $\bm{\Gamma}^{\text{c}}(\mathcal{N}; n)$. However, the advantage of defining the circularity indicators from $\bm{\Gamma}^{\text{c}}(\mathcal{N}; n)$, and hence, from a digraph, is in the generality of the approach. Indeed, the same graph-based procedure can be used to develop circularity indicators in continuous-time (e.g., see \cite{zocco2022circularity}), in discrete-time (e.g., in this paper), with smaller networks (e.g., the robotic cell covered here) and with larger networks such as the supply chains (e.g., see \cite{zocco2023thermodynamical}). 
\end{remark}

\begin{remark}
From Definition \ref{def:ThermoCircular}, it follows that the cell is \emph{not thermodynamically circular} since the cell digraph (Fig. \ref{fig:CellGraph}) has not any $\phi$, i.e., it has no closed loops. This matches with the reality since a robotic cell cannot close the material flow on its own; it can do so if it is part of a recovery chain that involves also the transportation and reuse of the materials or products.    
\end{remark}

\begin{remark}
The systemic framework leveraged in this paper enhances MFA as it is not merely based on mass balances, but it also considers the energy balances (i.e., the first principle of thermodynamics). Specifically, in this paper, the first principle of thermodynamics was covered in Section \ref{sub:RobotAsThermoComp} to derive the mechanics of robots, while the mass balances were used for deriving the circularity indicators in Section \ref{subsec:CircInd}.  
\end{remark}

\section{Results and Discussion}\label{sec:results}
This section covers the current capabilities of the flexible robotic cell along with discussions on future developments. Then, it illustrates a numerical example about the robotic cell circularity.

\subsection{Current Functionality of Robotic Cell}\label{subsec:cellPerformance}
\begin{figure*}
\begin{subfigure}{0.22\textwidth}
  \centering
  \includegraphics[width=.99\linewidth]{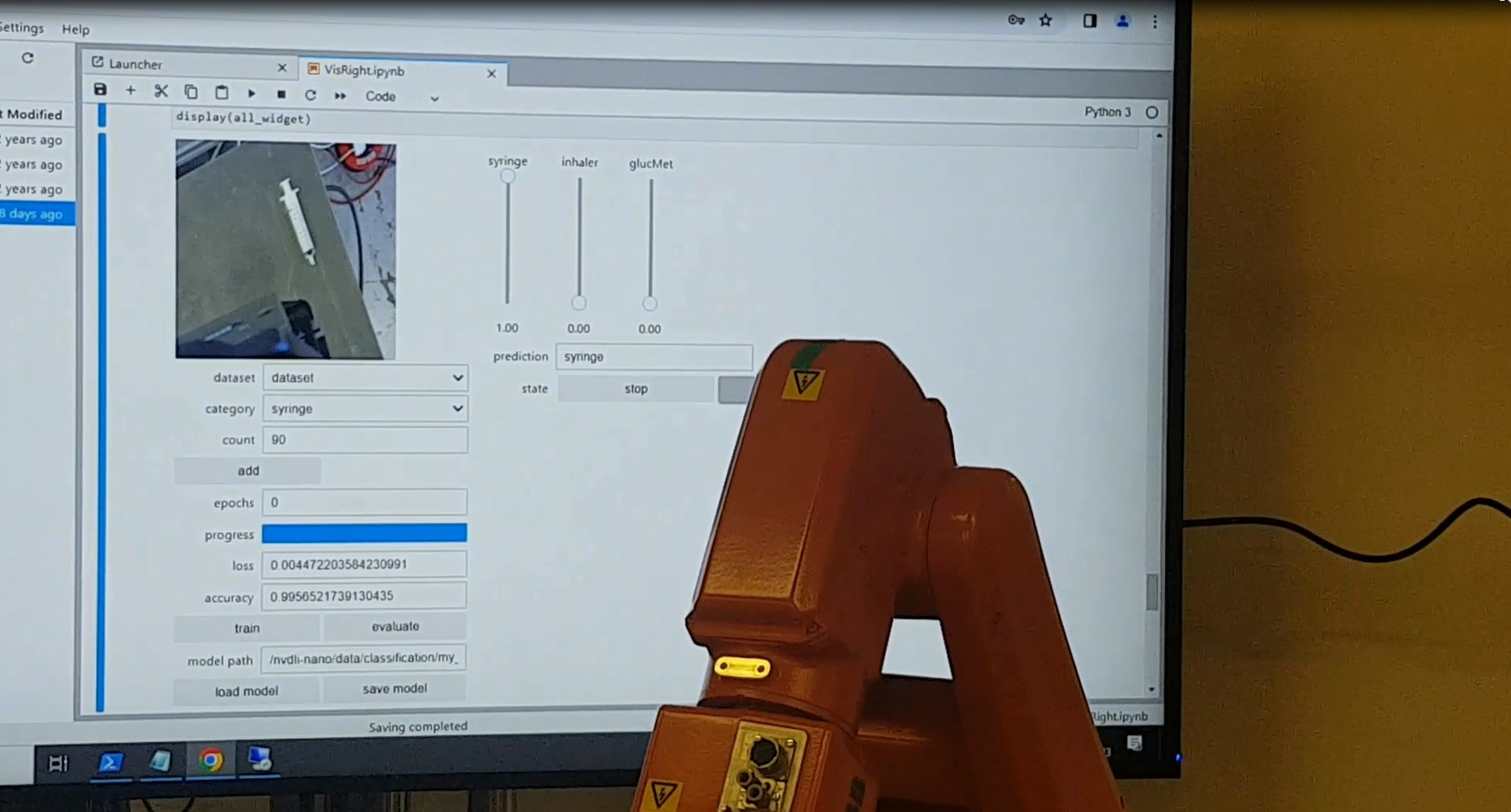}
  \caption{Real-time classification}
  \label{fig:RTclass}
\end{subfigure}%
\begin{subfigure}{0.27\textwidth}
  \centering
  \includegraphics[width=.99\linewidth]{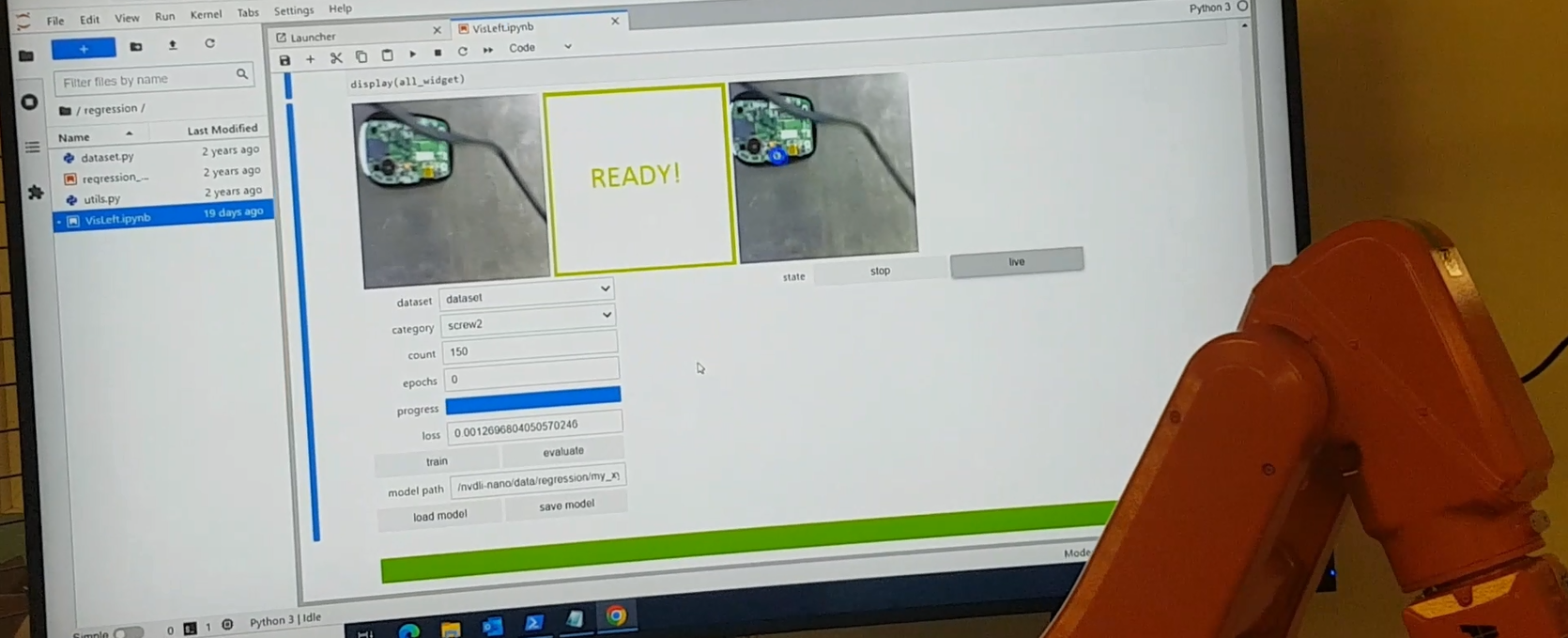}
  \caption{Real-time tracking of middle screw}
  \label{fig:RTdet1}
\end{subfigure}
\begin{subfigure}{0.25\textwidth}
  \centering
  \includegraphics[width=.99\linewidth]{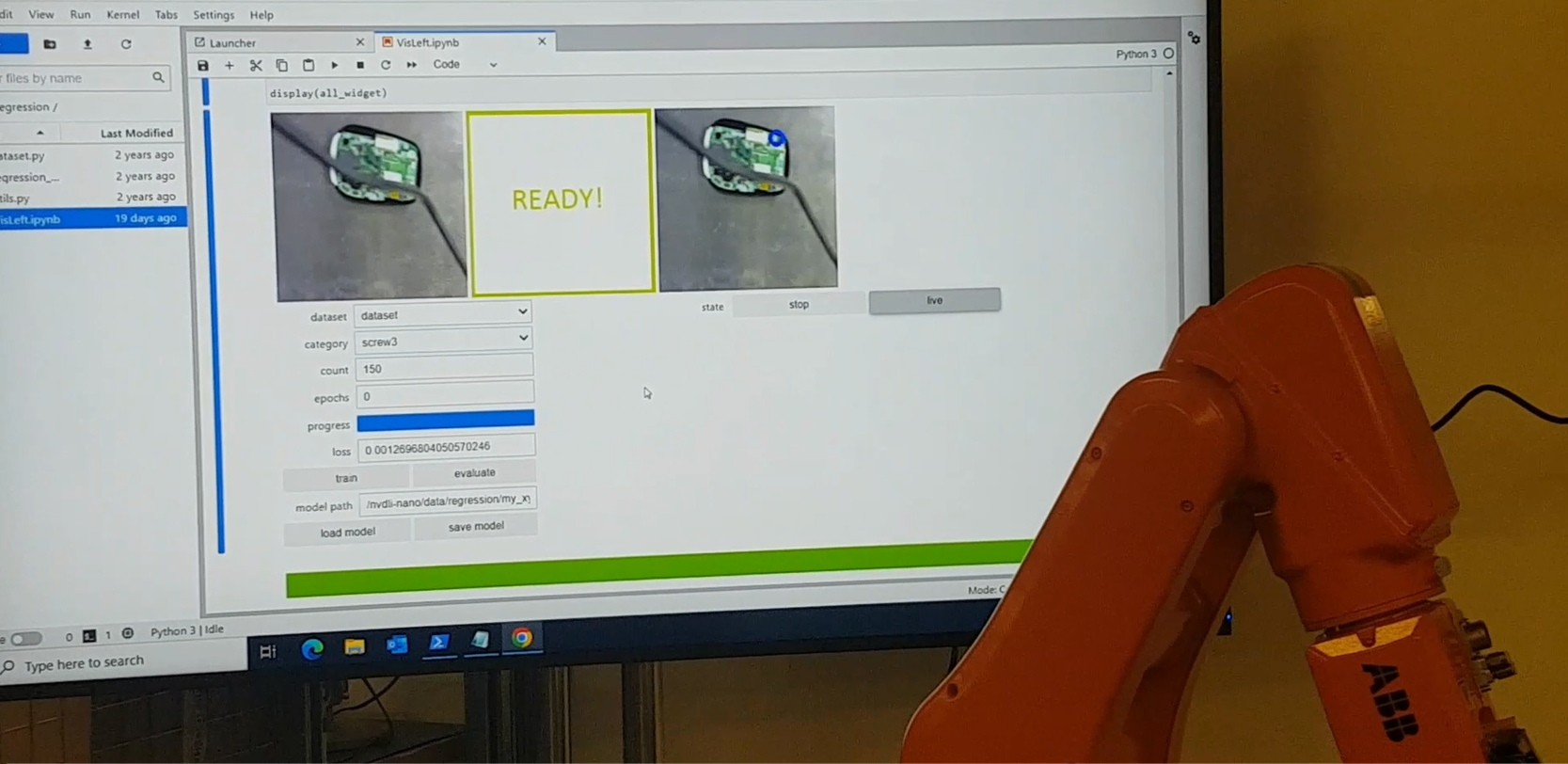}
  \caption{Real-time tracking of top-right screw}
  \label{fig:RTdet2}
\end{subfigure}
\begin{subfigure}{0.24\textwidth}
  \centering
  \includegraphics[width=.99\linewidth]{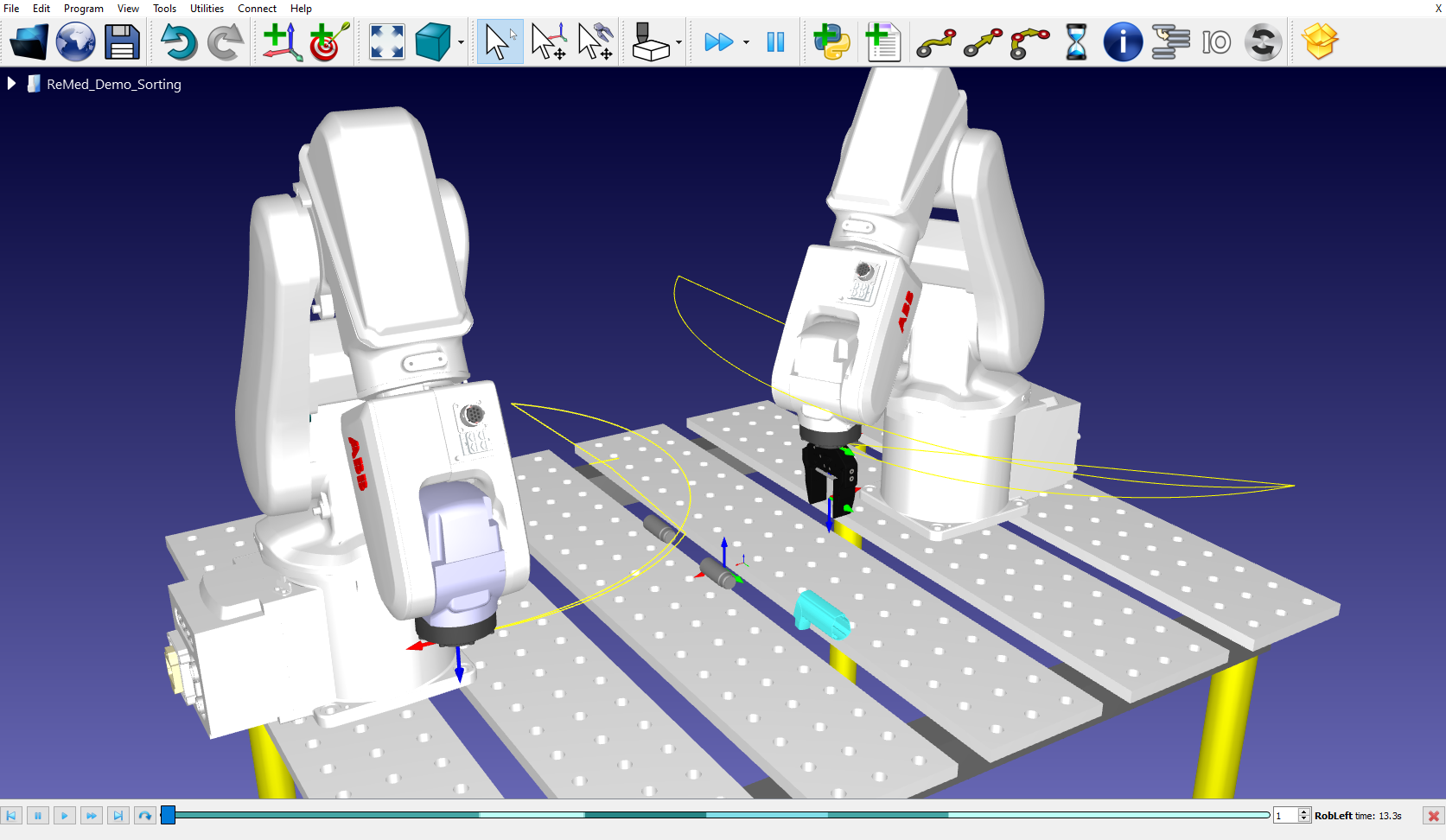}
  \caption{Simulator of trajectories}
  \label{fig:simulatorRes}
\end{subfigure}  
\caption{Current functionalities of the flexible robotic cell. Inhaler CAD model taken from \cite{GrabCAD}.}
\label{fig:results}
\end{figure*}

The current functionalities of the cell are depicted in Fig. \ref{fig:results}. 
Specifically, Fig. \ref{fig:RTclass} shows the camera view (top left) and the correct classification of a syringe indicated by the cursor (next to camera view). The neural model has 99\% accuracy on collected images (bottom left). Fig. \ref{fig:RTdet1} shows the real-time tracking of a screw indicated by the blue circle even with the disturbance of a cable, while Fig. \ref{fig:RTdet2} considers another screw. These screws secure the glucose meter PCB to the plastic case. The manipulators are programmed to move autonomously, the programs are exported from RoboDK and loaded into the IRC5 controllers. The RoboDK simulator in Fig. \ref{fig:simulatorRes} shows the robot trajectories in yellow to check the robot behavior before running the program into the real machine. See the demo video\footnotemark[1] for further details. 

Future work will enable communication between the microprocessors and the manipulators to synchronize the visual inference with the robot dynamics. We will also switch from classification to object detection to get the position of multiple items in the scene as required in realistic waste sorting scenarios \citep{zocco2023towards}. For performing autonomous resources mapping and quantification, we will connect the three webcams installed in the robotic cell to implement a first prototype of the system whose numerical study was covered in \cite{zocco2023visual} and initially proposed in \cite{zocco2022material}.

\subsection{Numerical Example on Cell Circularity}\label{subsec:exampleIndicators}
To illustrate the practical use of $\bm{\Gamma}^{\text{c}}$, $r_s$, and $t_s$, let us assume the following scenario: a GlucoRx HCT glucose meter has to be disassembled by a robotic cell; the masses of its parts are given in Table \ref{tab:NumExample}. 
\begin{table}
\centering
\caption{Masses of the parts of the considered glucose meter and the bin indices.}
\label{tab:NumExample}
\begin{tabular}{ccc} 
\hline
Part & Mass & $d$\\ 
\hline
Front case & 14.4 g & 1\\ % bin 1
Back case & 17.2 g & 1\\ % bin 1
PCB & 17.8 g & 2\\ % bin 2
Screw & 0.2 g & 3\\ % x 5 for bin 3
Spring & 0.1 g & 4\\ % everything else, bin 4
Button and clip & 1.6 g & 4\\ % everything else, bin 4
Test strip port & 1.0 g & 4\\ % everything else, bin 4 
Screen & 8.4 g & 4\\ % everything else, bin 4
USB port cap & 0.3 g & 4\\ % everything else, bin 4
\hline 
Whole ($m_{\text{gm}}$) & 61.8 g & - \\
\hline
\end{tabular}
\end{table}
The layout of the cell has 4 bins: the first bin is for the casing, the second bin is for the printed circuit board (PCB), the third bin is for the 5 screws, and the fourth bin is for any other parts. Hence, in this case, $D = 4$. Let us assume also that the situation is monitored every second, i.e., $T = 1$ s. The times at which the masses are moved are: the glucose meter enters the disassembly line at $n_{2,\text{in}} = 30$ after being extracted from a stock of 2 glucose meters at $n_1 = 5$; the casing, the PCB, the screws, and the remaining parts leave the disassembly line at $n_{2,\text{out},3} = 240$, $n_{2,\text{out},4} = 300$, $n_{2,\text{out},5} = 320$, and $n_{2,\text{out},6} = 360$, respectively. The casing, the PCB, the screws, and the remaining parts enter the designated bin at $n_{d,\text{in}} = n_{2,\text{out},d} + 5$, that is, the transfer from the disassembly line to the $d$-th bin takes 5 sample times for $d = 1, \dots, 4$. This transfer time is the time that it takes a robotic arm to pick a disassembled part (e.g., the PCB) and place it into the designated bin. Thus, the time for separating a part from the glucose meter is different from the time needed for moving the same part to the bin; the former is, in general, significantly longer than the latter.

With this scenario, the mass-flow matrix of the robotic cell (\ref{eq:GammaOfCell}) becomes
\begin{equation}\label{eq:GammaOfCellExplicit}
\bm{\Gamma}^{\text{c}}(\mathcal{N}; n) =
\begin{bmatrix}
\theta_0 & \theta_1 & 0 & 0 & 0 & 0 \\
0 & \theta_2 &  \theta_3 & \theta_4 & \theta_5 & \theta_6 \\
0 & 0 & \theta_7 & 0 & 0 & 0 \\
0 & 0 & 0 & \theta_8 & 0 & 0 \\
0 & 0 & 0 & 0 & \theta_9 & 0 \\
0 & 0 & 0 & 0 & 0 & \theta_{10} 
\end{bmatrix},
\end{equation}
where 
\begin{equation}
\theta_0 = m_1(n) - \overline{m}_{1,2} \delta_{n_1}(n),
\end{equation}
\begin{equation}
\theta_1 = m_{1,2}(n) + \overline{m}_{1,2} \left(\delta_{n_1}(n) - \delta_{n_{2,\text{in}}}(n)\right), 
\end{equation}
\begin{equation}
\begin{gathered}
\theta_2 = m_2(n) + \overline{m}_{1,2} \delta_{n_{2,\text{in}}}(n) - \overline{m}_{2,3} \delta_{n_{2,\text{out},3}}(n) \\ - \overline{m}_{2,4} \delta_{n_{2,\text{out},4}}(n) - \overline{m}_{2,5} \delta_{n_{2,\text{out},5}}(n) - \overline{m}_{2,6} \delta_{n_{2,\text{out},6}}(n), 
\end{gathered}
\end{equation}
\begin{equation}
\theta_3 = m_{2,3}(n) + \overline{m}_{2,3} \left(\delta_{n_{2,\text{out},3}} \right. \\ \left. - \delta_{n_{3,\text{in}}} \right), 
\end{equation}
\begin{equation}
\theta_4 = m_{2,4}(n) + \overline{m}_{2,4} \left(\delta_{n_{2,\text{out},4}} \right. \\ \left. - \delta_{n_{4,\text{in}}} \right),  
\end{equation}
\begin{equation}
\theta_5 = m_{2,5}(n) + \overline{m}_{2,5} \left(\delta_{n_{2,\text{out},5}} \right. \\ \left. - \delta_{n_{5,\text{in}}} \right),  
\end{equation}
\begin{equation}
\theta_6 = m_{2,6}(n) + \overline{m}_{2,6} \left(\delta_{n_{2,\text{out},6}} \right. \\ \left. - \delta_{n_{6,\text{in}}} \right),  
\end{equation}
\begin{equation}
\theta_7 = m_{3}(n) + \overline{m}_{2,3} \delta_{n_{2,\text{out},3}}(n),
\end{equation}
\begin{equation}
\theta_8 = m_{4}(n) + \overline{m}_{2,4} \delta_{n_{2,\text{out},4}}(n), 
\end{equation}
\begin{equation}
\theta_9 = m_{5}(n) + \overline{m}_{2,5} \delta_{n_{2,\text{out},5}}(n),
\end{equation}
and
\begin{equation}
\theta_{10} = m_{6}(n) + \overline{m}_{2,6} \delta_{n_{2,\text{out},6}}(n).
\end{equation}
Moreover, the circularity indicators of the cell become $t_s = (n_{2,\text{out},6} - n_{2,\text{in}})T = 330$ s and $r_s = 6$ (from (\ref{eq:sepTime}) and (\ref{eq:sepRate}), respectively). 

\begin{figure}[t!]
    \centering
    \begin{subfigure}[t]{0.24\textwidth}
        \centering
        \includegraphics[height=1.5in]{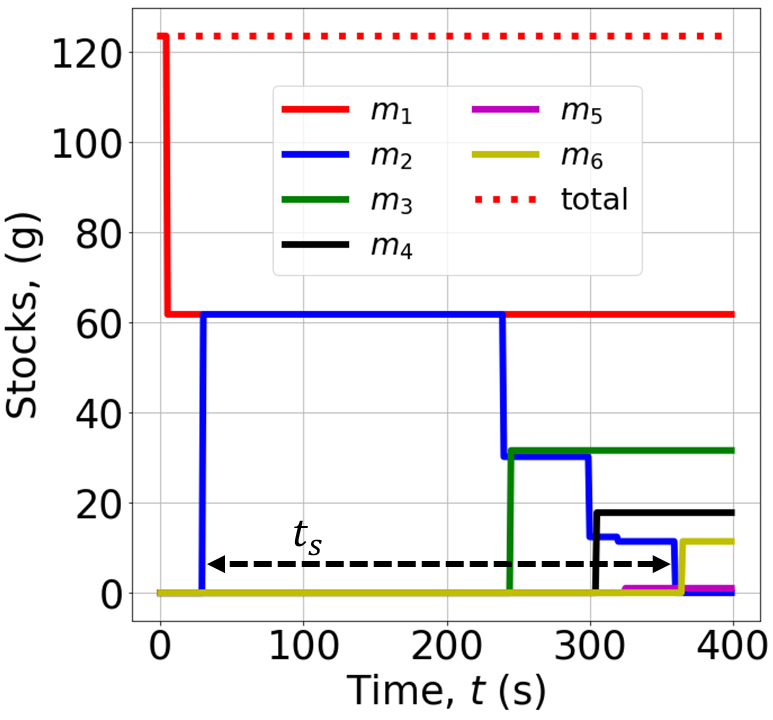}
        \caption{Stocks vs. time}
        \label{subfig:stocks}
    \end{subfigure}%
    ~ 
    \begin{subfigure}[t]{0.24\textwidth}
        \centering
        \includegraphics[height=1.5in]{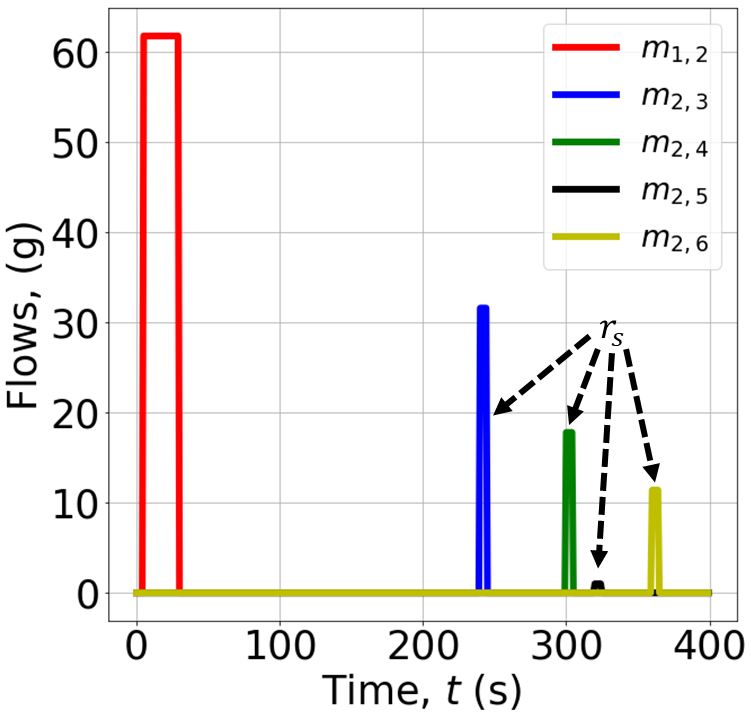}
        \caption{Flows vs. time}
        \label{subfig:flows}
    \end{subfigure}
    \caption{Dynamics of the stocks and flows of the robotic cell, which correspond to the entries $\theta_0, \theta_1, \dots, \theta_{10}$ of the mass-flow matrix (\ref{eq:GammaOfCellExplicit}).}
\label{fig:simulationResults}
\end{figure}

Now, let us analyze the dynamics of stocks and flows within the system, which are shown in Fig. \ref{fig:simulationResults}. Then, we will analyze how they affect the circularity indicators $t_s$ and $r_s$. As visible in Fig. \ref{subfig:stocks}, $m_1$ is the only stock for $t < n_1 = 5$ s, whose value is $m_1 = 2m_{\text{gm}} = 123.6$ g. Then, as visible in Fig. \ref{subfig:flows}, $m_{1,2} = m_{\text{gm}} = 61.8$ g for $n_1 \leq t < n_{2,\text{in}}$. For $n_{2,\text{in}} \leq t < n_{2,\text{out},3}$, it holds that $m_2 = m_{\text{gm}}$. In particular, the dynamics of $m_2$ has 5 variations: the first variation is in $t = n_{2,\text{in}}$ and it is an increase of mass because $m_{\text{gm}}$ enters, whereas the other 4 variations are reductions because the mass exits to enter the 4 designated bins. These 4 mass transfers can be seen easily in Fig. \ref{subfig:flows} as they correspond to the 4 peaks in $m_{2,3}$, $m_{2,4}$, $m_{2,5}$, and $m_{2,6}$. These 4 peaks are rectangles with duration of 5 s, which is the transfer time of the glucose meter parts from the disassembly line ($m_2$) to any bin. The tallest rectangle in Fig. \ref{subfig:flows} is $m_{1,2}$, which is the transfer of the glucose meter (61.8 g) from the stock $m_1$ to the disassembly line $m_2$. Note also from Fig. \ref{subfig:stocks} that $m_3$, $m_4$, $m_5$, and $m_6$ have an increase with the same magnitude and immediately after the corresponding decrease of $m_2$ since the disassembled parts exit $m_2$ to enter $m_3$, $m_4$, $m_5$, and $m_6$. The third bin (i.e., $m_5$) receives the 5 screws, hence it contains the smallest mass, that is, $5 \times 0.2 = 1$ g. As the whole robotic cell is a closed system, the total mass within it is constant as showed in Fig. \ref{subfig:stocks} and it is equal to $2 \times m_{\text{gm}} = 123.6$ g.                     

The performance of the robotic cell in terms of circularity is measured by the indicators $t_s$ (\ref{eq:sepTime}) and $r_s$ (\ref{eq:sepRate}). The former is indicated in Fig. \ref{subfig:stocks} and it is affected by the speed of the disassembly, which can be increased, for example, by speeding-up the disassembly sequence or by reducing the image processing latency. The latter is indicated in Fig. \ref{subfig:flows} and, in contrast with $t_s$, an higher value leads to a better cell from the circularity perspective. Indeed, a cell with higher $r_s$ can separate more materials to facilitate their reuse or recycling. Intuitively, $r_s$ can be increased by increasing the number of bins $D$ since $r_s = 2+D$. However, this may lead to a more complex disassembly sequence, and hence, it may increase $t_s$ (e.g., with $D = 1$ no disassembly is performed since the product goes straight into the only bin; in this case, $r_s = 3$, which is the worst value, whereas $t_s = 0$ s, which is ideal). Therefore, improving $r_s$ and $t_s$ jointly can be particularly challenging. In this case, $t_s = 330$ s and $r_s = 6$.

\section{Conclusions}\label{sec:concl}
This paper presented the development of a flexible robotic cell for reprocessing small medical devices without contamination risks and also proposed two indicators to measure the robotic cell circularity leveraging a thermodynamics-based systemic modeling framework. 

% situation of the cell and main future work
As of January 2024, the phase of building the robotic cell is almost complete, whereas the phase of programming has yet to reach its peak. Indeed, as soon as the communication between the microprocessors and the manipulators is established, the visual inference should provide a real-time and accurate feedback to the robot controllers to achieve a human-like adaptive behavior. A key challenge will be to find a compromise between the visual feedback accuracy, the computation latency, and the communication latency to effectively perform waste sorting and disassembly operations of items located in random positions.    

% integration of robotics into systemic
The integration of robot dynamics into a systemic modeling framework enables us to embed the modeling and control of industrial manipulators into the design of recovery chains, and thus, to regulate the robot performance with respect to the material flow circularity indicators.  

% evaluation of circularity indicators  
The numerical study on circularity indicators highlighted that the separation time, that is, the robot speed, is a key factor as in the manufacturing processes of linear economies. However, differently from manufacturing, the rate of material separation, i.e., $r_s$, is a key factor for reprocessing operations since it facilitates the reuse of parts, the repair of products, and the material recycling. 

% Step forward in CE foundations:
This paper also has coherently integrated deep-learning vision, robotics theory, thermodynamics, and graph theory to make a step forward in the definition of the theoretical foundations of circular economy as a scientific discipline. Indeed, while the behavioral principles for improving circularity are clear, i.e., reduce, reuse, repair, recycle, etc., the actual design of circular material flows is currently lacking the mathematical rigor existing in other more mature fields such as electrical network design \citep{Balabanian1969}, robot design \citep{SicilianoBook}, vehicle design \citep{guiggiani2014science}, and artificial neural network design \citep{goodfellow2016deep}.

\section*{Acknowledgements}
This work has been conducted as part of the research project ‘Circular Economy for Small Medical Devices (ReMed)’, which is funded by the Engineering and Physical Sciences Research Council (EPSRC) of the UKRI (contract no: EP/W002566/1). The project funders were not directly involved in this work. 
The authors gratefully thank the ReMed Team (\url{https://www.remed.uk/team}) and Dan Lake at Loughborough University for the valuable discussions.

\bibliographystyle{elsarticle-harv}
\bibliography{references}        

\end{document}